\documentclass[letterpaper, 10 pt, conference]{ieeeconf}  % Comment this line out if you need a4paper
\IEEEoverridecommandlockouts                              % This command is only needed if 
\usepackage[noadjust]{cite}
\usepackage{graphicx}
\usepackage{amsmath}
\usepackage{amssymb}
\usepackage{amsfonts}
\usepackage{mathtools, xparse}
\usepackage{algorithm}
\usepackage{algorithmicx}
\usepackage{algpseudocode}

\usepackage{multirow}
\usepackage[normalem]{ulem}
\useunder{\uline}{\ul}{}
\usepackage{hyperref}

\newcommand{\URL}{www.goo.gl/b57WTs}
\title{\LARGE \bf Asymmetric Actor Critic for Image-Based Robot Learning}
\author{Lerrel Pinto$^{1,2}$  Marcin Andrychowicz$^{1}$ Peter Welinder$^{1}$ Wojciech Zaremba$^{1,\dagger}$ Pieter Abbeel$^{1,\dagger}$\\
\thanks{$^{1}$OpenAI {\tt\small \{marcin, pw, woj, pieter\}@openai.com}}%
\thanks{$^{2}$Carnegie Mellon University {\tt\small lerrelp@cs.cmu.edu}}%
\thanks{$^{\dagger}$Equal advising}
}
\begin{document}
\maketitle
\thispagestyle{empty}
\pagestyle{empty}
\begin{abstract}
Deep reinforcement learning (RL) has proven a powerful technique in many sequential decision making domains. However, Robotics poses many challenges for RL, most notably training on a physical system can be expensive and dangerous, which has sparked significant interest in learning control policies using a physics simulator. While several recent works have shown promising results in transferring policies trained in simulation to the real world, they often do not fully utilize the advantage of working with a simulator. In this work, we exploit the full state observability in the simulator to train better policies which take as input only partial observations (RGBD images). We do this by employing an \textit{actor-critic} training algorithm in which the critic is trained on full states while the actor (or policy) gets rendered images as input. We show experimentally on a range of simulated tasks that using these asymmetric inputs significantly improves performance. Finally, we combine this method with domain randomization and show real robot experiments for several tasks like picking, pushing, and moving a block. We achieve this simulation to real world transfer without training on any real world data. Videos of these experiments can be found at \url{\URL}.

\end{abstract}
\section{INTRODUCTION}
Reinforcement learning (RL) coupled with deep neural networks has recently led to successes on a wide range of control problems, including achieving superhuman performance on Atari games~\cite{mnih2015human} and beating the world champion in the classic game of Go~\cite{silver2016mastering}. In physics simulators, complex behaviours like walking, running, hopping and jumping have also been shown to emerge~\cite{schulman2015trust,lillicrap2015continuous}.

In the context of robotics however, learning complex behaviours faces two unique challenges: scalability and safety. Robots are slow and expensive which makes existing data intensive learning algorithms hard to scale. These physical robots could also damage themselves and their environment while exploring these behaviours. A recent approach to circumvent these challenges is to train on a simulated version of the robot and then transfer to the real robot~\cite{cutler2015efficient, ghadirzadeh2017deep, kolter2007learning, rusu2016sim,christiano2016transfer,zhang2016vision,sadeghi2016cad,held2017probabilistically}. However, this brings about a new challenge: observability.

\begin{figure}[h]
\begin{center}
\includegraphics[width=3.2in]{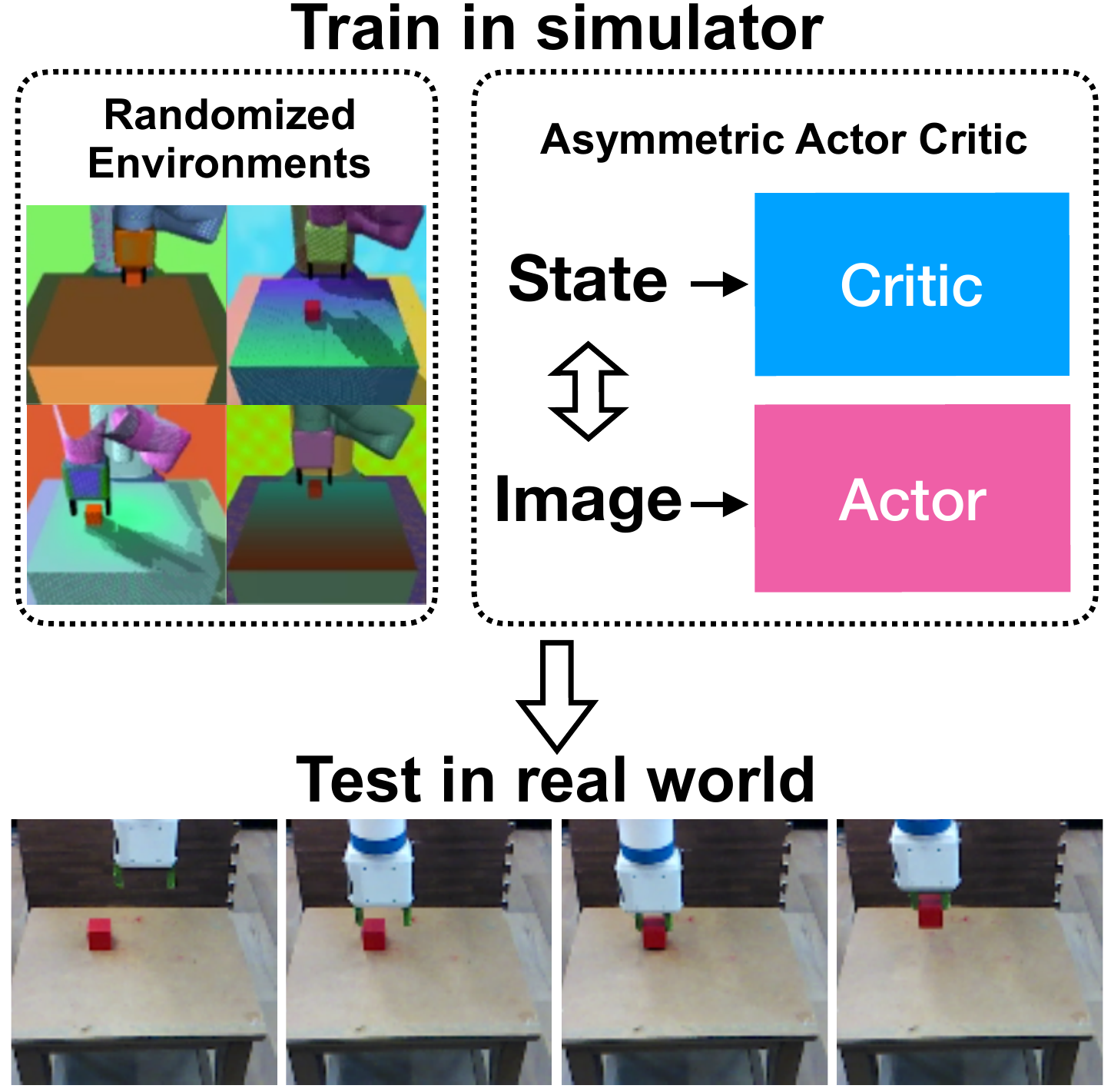}
\end{center}
\caption{By training policies with asymmetric inputs for actor-critic along with domain randomization, we learn complex visual policies that can operate in the real without having seen any real world data in training.}
\label{fig:intro}
\end{figure}

Simulators have access to the full state of the robot and its surroundings, while in the real world obtaining this full state observability is often infeasible. One option is to infer the full state by visual detectors ~\cite{girshick2014rich, andrychowicz2017hindsight} or state prediction filters ~\cite{kalman1960new}. Explicit full state prediction from partial observations is often impossible and this challenge is further exacerbated by the compounding error problem ~\cite{venkatraman2016improved}. Another option is to train entirely on rendered partial observations (camera images) of the robot ~\cite{tobin2017domain, sadeghi2016cad}. However, these techniques are not powerful enough to learn complex behaviours due to the large input dimensionality and partial observability. This leads us to a conundrum, i.e., training on full states is hard since it depends on good state predictors while training on images is hard because of their partial observability and dimensionality. We solve this by learning a policy that relies only on partial observations (RGBD images) but during training we exploit access to the full state.

Physics simulators give us access to both the full state of the system as well as rendered images of the scene. But how can we combine these observations to train complex behaviours faster? In this work, we exploit this access and train an \textit{actor-critic} algorithm~\cite{konda2000actor,lillicrap2015continuous} that uses asymmetric inputs, i.e. the actor takes visual partial observations as input while the critic takes the underlying full state as input. 
% Since training is performed in a simulator, we use the full state to train the critic while using the rendered visual observations as input for the actor. %%PA: last sentence is too repetitive wrt to previous sentence, maybe remove (or merge into previous whatever info wasn't there yet)
Since the critic works on full state, it learns the \textit{state-action} value function much faster, which also allows for better updates for the actor. During testing, the actor is employed on the partial observations and does not depend on the full state (the full state is only used during training). This allows us to train an actor/policy network on visual observations while exploiting the availability of full states to train the critic. We experimentally show significant improvements on 2D environments like \textit{Particle} and \textit{Reacher} and 3D environments like \textit{Fetch Pick}. To further speed up training, we also demonstrate the utility of using bottlenecks ~\cite{zhang2016vision}.

Another key aspect of this work is to show that these policies learned in a simulator can be transferred to the real robot without additional real world data. Simulators are not perfect representations of the real world. The domain of observations (real camera images) significantly differs from rendered images from a simulator. This makes directly transferring policies from the simulator to the real world hard. However, ~\cite{tobin2017domain, viereck2017learning, sadeghi2016cad} show how randomizing textures and lighting allows for effective transfer. By combining our asymmetric actor critic training with domain randomization ~\cite{tobin2017domain}, we show that these policies can be transferred to a real robot without any training on the physical system (see Figure~\ref{fig:intro}).

\section{RELATED WORK}

\subsection{Reinforcement Learning}
Recent works in deep reinforcement learning (RL) have shown impressive results in the domain of games~\cite{mnih2015human,silver2016mastering} and simulated control tasks~\cite{schulman2015trust,lillicrap2015continuous}. The class of RL algorithms our method employs are called \textit{actor-critic} algorithms~\cite{konda2000actor}. Deep Deterministic Policy Gradients (DDPG)~\cite{lillicrap2015continuous} is a popular \textit{actor-critic} algorithm that has shown impressive results in continuous control tasks. Although we use DDPG for our base optimizer, our method is applicable to arbitrary \textit{actor-critic} algorithms.

Learning policies in an environment that provides only sparse rewards is a challenging problem due to very limited feedback signal. However it has been shown that sparse rewards often allow for better policies when trained appropriately~\cite{andrychowicz2017hindsight}. Moreover having sparse rewards allows us to circumvent manual shaping of the reward function.

The idea of using different inputs for the actor and critic has been explored previously in the domain of multi-agent learning ~\cite{foerster2017counterfactual,lowe2017multi}. 
However using this in the domain of robot learning and dealing with partial observability hasn't been explored. Exploiting the access to full state in training the critic also draws similarities to Guided Policy Search~\cite{levine2016end}.

\subsection{Transfer from simulation to the real world}
Bridging the reality gap in transferring policies trained in a simulator to the real world is an active area of research in the robot learning community. One approach is to make the simulator as close to the real world as possible ~\cite{james20163d,planche2017depthsynth,richter2016playing}. But these methods have had limited success due to the hard system identification problem. 

Another approach is domain adaptation from the simulator~\cite{cutler2015efficient, ghadirzadeh2017deep, kolter2007learning, rusu2016sim,christiano2016transfer,zhang2016vision}, since it may be easier to finetune from a simulator policy than training in the real world. However if the simulator differs from the real world by a large factor, the policy trained in simulation can perform very poorly in the real world and finetuning may not be any easier than training from scratch. This limits most of these works to learning simple behaviours. Making policies robust for physics adaptation~\cite{rajeswaran2016epopt, pinto2017robust, yu2017preparing} is also receiving interest, but these methods haven't been shown to be powerful enough to work on real robots. Using bottlenecks~\cite{zhang2016vision} has been shown to help domain adaptation for simple tasks like \textit{reaching}. In this work, we show how bottlenecks can be exploited for more complex fine manipulation tasks. 

A promising approach is domain adaptation by domain randomization ~\cite{tobin2017domain,sadeghi2016cad}. Here the key idea is to train on randomized renderings of the scene, which allows to learn robust policies for transfer. However these works do not show transfer to precise manipulation behaviours. We show that this idea can be extended to complex behaviours when coupled with our asymmetric actor critic.

\subsection{Robotic tasks}
We perform real robot experiments on tasks like picking, pushing, and moving a block. The \textit{Picking} task is similar to grasping objects~\cite{bicchi2000robotic,pinto2016supersizing}, however in this work we learn an end-to-end policy that moves to the object, grasps it and moves the grasped object to its desired position. The focus is hence on the fine manipulation behaviour. The tasks of \textit{Forward Pushing} and \textit{Block Move} are similar to pushing objects~\cite{balorda1990reducing,pinto2017learning}, however as in the case of \textit{Picking}, this paper focuses on the learning of the fine pushing behaviour.
% -Methods for sim2real. cad2rl, progressivenets, baxterreacher, domainrandomization, 
% -Methods of randomization.
% -Methods for robustness. EPopt, RARL, physics adaptation
% -Methods for bottleneck

\section{BACKGROUND}
Before we discuss our method, we briefly introduce some background and formalism for the RL algorithms used. A more comprehensive introduction can be found in ~\cite{kaelbling1996reinforcement}.

\subsection{Reinforcement Learning}
In this paper we deal with continuous space Markov Decision Processes that can be represented as the tuple $(\mathcal{S},\mathcal{O},\mathcal{A},\mathcal{P},r,\gamma, \mathbb{S})$, where $\mathcal{S}$ is a set of continuous states and $\mathcal{A}$ is a set of continuous actions, $\mathcal{P}: \mathcal{S} \times \mathcal{A} \times \mathcal{S} \rightarrow \mathbb{R}$ is the transition probability function, $r: \mathcal{S} \times \mathcal{A} \rightarrow \mathbb{R}$ is the reward function, $\gamma$ is the discount factor, and $\mathbb{S}$ is the initial state distribution. $\mathcal{O}$ is a set of continuous partial observations corresponding to states in $\mathcal{S}$.

An episode for the agent begins with sampling $s_0$ from the initial state distribution $\mathbb{S}$. 
%%PA: usually s0 denotes the state at time 0, not the initial state distribution; probably good to stick with convention
At every timestep $t$, the agent takes an action $a_t=\pi(s_t)$ according to a deterministic policy $\pi:\mathcal{S} \rightarrow \mathcal{A}$. At every timestep $t$, the agent gets a reward $r_t=r(s_t,a_t)$, and the state transistions to $s_{t+1}$, which is sampled accordingly to probabilities $\mathcal{P}(s_t,a_t,\cdot)$. The goal of the agent is to maximize the expected return $E_{\mathbb{S}}[R_0|\mathbb{S}]$, where the return is the discounted sum of the future rewards $R_t=\sum^{\infty}_{i=t}\gamma^{i-t}r_i$. The $Q$-function is defined as $Q^{\pi}(s_t,a_t)=E[R_t|s_t,a_t]$. In the partial observability case, the agent takes actions based on the partial observation, $a_t=\pi(o_t)$, where $o_t$ is the observation corresponding to the full state $s_t$. 
%%PA: do we consider memory in this work? i.e. might the action be based on  o_0 .. o_t ?  or just o_t?
%%LP: Nope. Policy could be recurrent.

\subsection{Deep Deterministic Policy Gradients (DDPG)}
Deep Deterministic Policy Gradients (DDPG)~\cite{lillicrap2015continuous} is an \textit{actor-critic} RL algorithm that learns a deterministic continuous action policy. The algorithm maintains two neural networks: the policy (also called the actor) $\pi_\theta:\mathcal{S} \rightarrow \mathcal{A}$ (with neural network parameters $\theta$) and a $Q$-function approximator (also called the critic) $Q_\phi^{\pi}:\mathcal{S} \times \mathcal{A} \rightarrow \mathbb{R}$ (with neural network parameters $\phi$). 

%%PA: might it be clearer to explicitly parameterize the policy and the Q function, e.g. \pi_\theta and Q_\phi ?  that'll make it more obvious what's being learnt?

During training, episodes are generated using a noisy version of the policy (called behavioural policy), e.g. $\pi_b(s) = \pi(s) + \mathcal{N}(0,1)$, where $\mathcal{N}$ is Normal noise. The transition tuples $(s_t,a_t,r_t,s_{t+1})$ encountered during training are stored in a replay buffer~\cite{mnih2015human}. Training examples sampled from the replay buffer are used to optimize the critic. By minimizing the Bellman error loss $\mathcal{L}_c=(Q(s_t,a_t)-y_t)^2$, where $y_t=r_t + \gamma Q(s_{t+1},\pi(s_{t+1}))$, the critic is optimized to approximate the $Q$-function. The actor is optimized by minimizing the loss $\mathcal{L}_a=-E_s[Q(s,\pi(s)]$. The gradient of $\mathcal{L}_a$ with respect to the actor parameters is called the deterministic policy gradient ~\cite{silver2014deterministic} and can be computed by backpropagating through the combined critic and actor networks. 

To stabilize the training, the targets for the actor and the critic $y_t$ are computed on separate versions of the actor and critic networks, 
%%PA: just the critic network?
which change at a slower rate than the main networks. A common practice is to use a Polyak averaged ~\cite{polyak1992acceleration} version of the main network.

\subsection{Multigoal RL}
We are interested in learning policies that can achieve multiple goals (a universal policy). One way of doing this is by training policies and $Q$-functions that take as an additional input a goal $g\in \mathcal{G}$ ~\cite{schaul2015universal,andrychowicz2017hindsight}, e.g. $a_t=\pi(s_t, g)$. A universal policy can hence be trained by using arbitrary RL algorithms. 

Following UVFA~\cite{schaul2015universal}, the sparse reward formulation $r(s_t,a,g)=[d(s_t,g)<\epsilon]$ will be used in this work, where the agent gets a positive reward when the distance $d(.,.)$ between the current state and the goal is less than $\epsilon$. In the context of a robot performing the task of picking and placing an object, this means that the robot gets a reward only if the object is within $\epsilon$ euclidean distance of the desired goal location of the object. Having a sparse reward overcomes the limitation of hand engineering the reward function, which often requires extensive domain knowledge. However, sparse rewards are not very informative and makes it hard to optimize. In order to overcome the difficulties with sparse rewards, we employ a recent method: Hindsight Experience Replay (HER)~\cite{andrychowicz2017hindsight}.

\subsection{Hindsight Experience Replay (HER)}
HER ~\cite{andrychowicz2017hindsight} is a simple method of manipulating the replay buffer used in off-policy RL algorithms that allows it to learn universal policies more efficiently with sparse rewards. After experiencing some episode $s_0,s_1,...,s_T$, every transition $s_t\rightarrow s_{t+1}$ along with the goal for this episode is usually stored in the replay buffer. However with HER, the experienced transitions are also stored in the replay buffer with different goals. These additional goals are states that were achieved later in the episode. Since the goal being pursued does not influence the environment dynamics, we can replay each trajectory using arbitrary goals assuming we use an off-policy RL algorithm to optimize~\cite{precup2001off}.

\section{METHOD}
We now describe our method along with the technique of bottlenecks to speed up training. Following this, we also describe domain randomization for transferring simulator learned policies to the real robot.

\begin{figure}[!h]
\begin{center}
\includegraphics[width=3.0in]{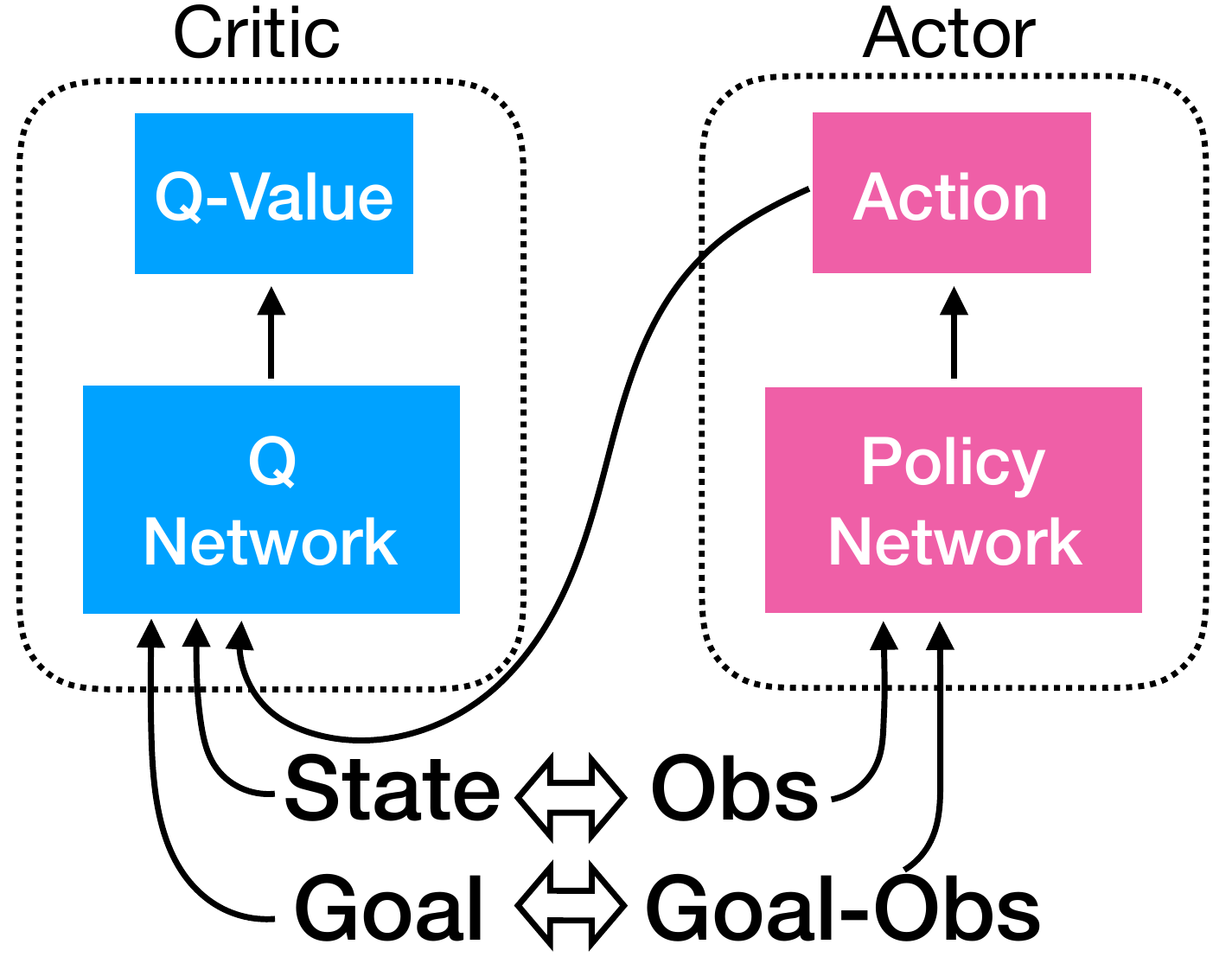}
\end{center}
\caption{Having asymmetric inputs, i.e. full states for the critic and partial observations for the actor improves training. In the multi goal setting, the critic additionally requires full goal states while the actor additionally requires partial observations for the goal.}
\label{fig:asym_ac}
\end{figure}

\subsection{Asymmetric Actor Critic}
In its essence our method builds on \textit{actor-critic} algorithms~\cite{konda2000actor} by using the full state $s_t \in \mathcal{S}$ to train the critic, while using partial observation $o_t \in \mathcal{O}$ to train the actor (see Figure ~\ref{fig:asym_ac}). Note that $s_t$ is the underlying full state for the observation $o_t$. In our experiments, observations $o_t$ are images taken by an external camera.

\begin{algorithm}
\begin{algorithmic}
% \State \textbf{Given:}
% \State an offpolicy actor critic algorithm $\mathcal{A}$
% \State 
\State Initialize actor-critic algorithm $\mathcal{A}$
\State Initialize replay buffer $R$
\For{episode$=1,M$} 
    \State Sample a goal $g$ and an initial state $s_0$
    \State Render goal observation $g^{o}$
    \State {\hspace{0.5in} $g^{o} \leftarrow \texttt{renderer}(g)$}
    \For{$t=0,T-1$}
        \State Render image observation $o_t$
        \State {\hspace{0.5in} $o_t \leftarrow \texttt{renderer}(s_t)$}
        \State Obtain action $a_t$ using behavioural policy:
        \State {\hspace{0.5in} $a_t \leftarrow \pi_b(o_t,g^{o})$}
        \State Execute action $a_t$, receive reward $r_t$ and transition to $s_{t+1}$
        \State Store $(s_t, o_t, a_t, r_t, s_{t+1}, o_{t+1}, g, g^{o})$ in $R$
    \EndFor
    %\For{$t=0,T-1$}
    %    \State $r_t := r(s_t,a_t,g)$
    %    \State Store transition $(s_t, o_t, a_t, r_t, s_{t+1}, o_{t+1}, g)$ in $R$
        % \State Sample additional goals $G:=\mathbb{S}(\textbf{current episode})$
        % \For {$g^{'}\in G$}
        %     \State $r^{'}_t:=r(s_t,a_t,g^{'})$
        %     \State Store $(s_t, o_t, a_t, r^{'}_t, s_{t+1}, o_{t+1}, g^{'})$ in $R$
        % \EndFor
    %\EndFor
    \For {n=1,\,N}
        \State Sample minibatch $\{s, o, a, r, s^{'}, o^{'}, g, g^{o}\}^{B}_0$ from $R$
        \State Optimize critic using $\{s, a, r, s^{'}, g\}^{B}_0$ with $\mathcal{A}$
        \State Optimize actor using $\{o, a, r, o^{'}, g^{o}\}^{B}_0$ with $\mathcal{A}$
    \EndFor
\EndFor
\end{algorithmic}
\caption{Asymmetric Actor Critic}
\label{alg:asym_her}
\end{algorithm}

The algorithm (described in Algorithm \ref{alg:asym_her}), begins with initializing the networks for an off-policy actor-critic algorithm $\mathcal{A}$ ~\cite{precup2001off}. In this paper, we use DDPG~\cite{lillicrap2015continuous} as the actor-critic algorithm. The replay buffer $R$ used by this algorithm is initialized with no data. For each episode, a goal $g$ and an initial state $s_0$ are sampled before the rollout begins. $g^{o}$ is the rendered goal observation. At every timestep $t$ of the episode, a partially observable image of the scene $o_t$ is rendered from the simulator at the full state $s_t$. The behavioural policy from $\mathcal{A}$, which is usually a noisy version of the actor is used to generate the action $a_t$ for the agent/robot to take. After taking this action, the environment transitions to the next state $s_{t+1}$, with its corresponding rendered image $o_{t+1}$. 

Since DDPG relies on a replay buffer to sample training data, we build the replay buffer from the episodic experience $(s_t, o_t, a_t, r_t, s_{t+1}, o_{t+1}, g, g^{o})$ previously generated. To improve performance for the sparse reward case, we augment the standard replay buffer by adding hindsight experiences ~\cite{andrychowicz2017hindsight}.

% by sampling additional goals $g^{'}$ according to a goal sampling strategy $\mathbb{S}$. A simple strategy for $\mathbb{S}$ is to select the last state encountered in the episode $s_T$ as the additional goal.

After the episodic experience has been added to the replay buffer $R$, we can now train our actor-critic algorithm $\mathcal{A}$ from sampled minibatch of size $B$ from $R$. This minibatch can be represented as $\{s, o, a, r, s^{'}, o^{'}, g, g^{o}\}^{B}_0$, where $s^{'}$ and $o^{'}$ are the next step full state and next step observation respectively. Since the critic takes full states as input, it is trained on $\{s, a, r, s^{'}, g\}^{B}_0$. Since the actor takes observations as input, it is trained on $\{o, a, r, o^{'}, g^{o}\}^{B}_0$. We experimentally show that asymmetric inputs for the critic and actor significantly improves performance and allows to transfer more complex manipulation behaviours to real robots.

\subsection{Improvements with bottlenecks}
One way of improving the efficiency of training is to use bottlenecks~\cite{zhang2016vision}. The key idea is to constrain one of the actor network's intermediary layers to predict the full state. Since the full state is often of a smaller dimension than the other layers of the network, this state predictive layer is called the bottleneck layer. 

\subsection{Randomization for transfer}

A powerful technique for domain transfer of policies from rendered images to real world images is domain randomization ~\cite{tobin2017domain, sadeghi2016cad}. The key idea is to randomize visual elements in the scene during the rendering. Learning policies with this randomization allows the policy to generalize to sources of error in the real world and latch on to the important aspects of the observation. 

\begin{figure}[!h]
\begin{center}
\includegraphics[width=3.3in]{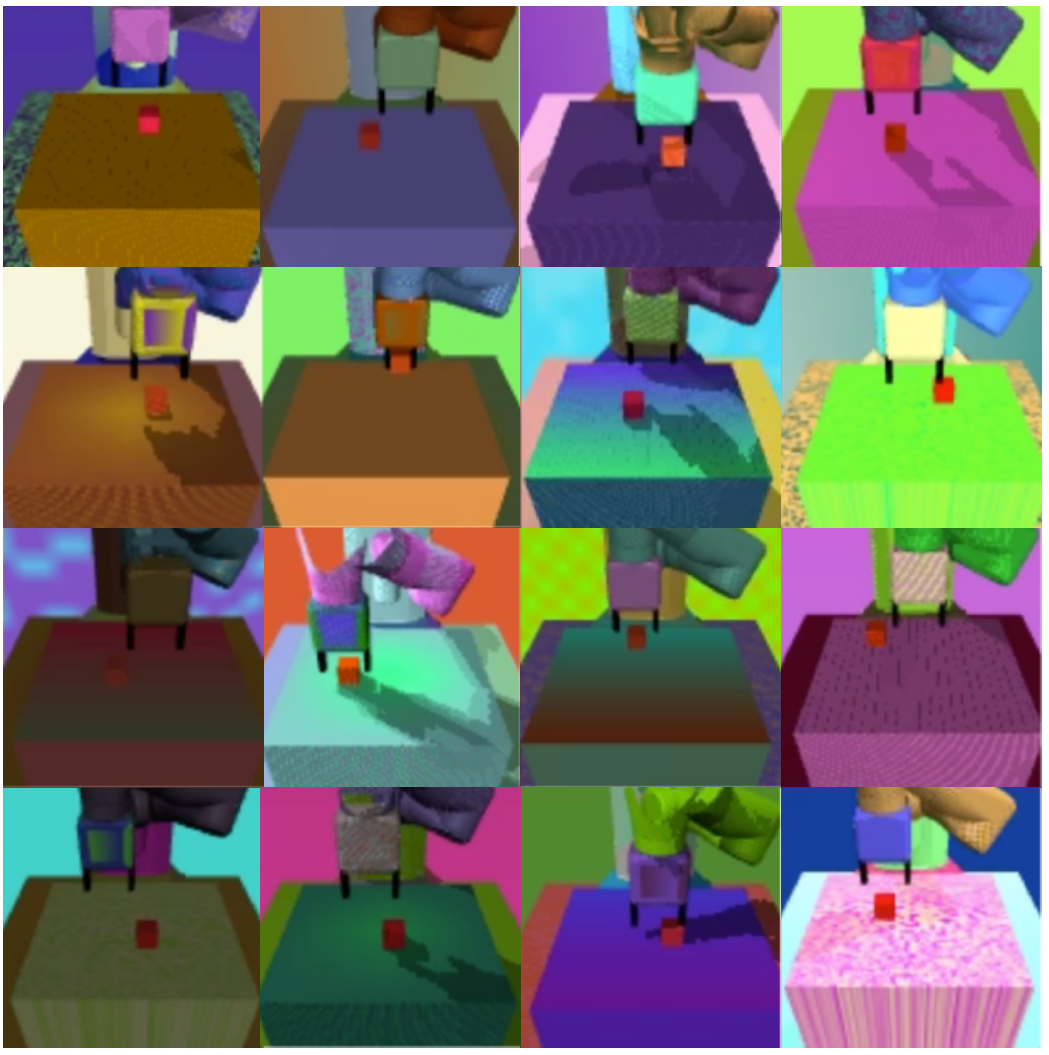}
\end{center}
\caption{To enable transfer of policies from the simulator to the real world, we randomize various aspects of the renderer during training. These aspects include textures, lighting and the position of the camera.}
\label{fig:randomness}
\end{figure}

For the purposes of this paper, we randomize the following aspects: texture, lighting, camera location and depth. For textures, random textures are chosen among random RGB values, gradient textures and checker patterns. These random textures are applied on the different physical objects in the scene, like the robot and the table. For lighting randomization, we randomly switch on lighting sources in the scene and also randomize the position, orientation and the specular characteristics of the light. For camera location, we randomize the location of the monocular camera in a box around the expected location of the real world camera. Furthermore, we randomize the orientation and focal length of the camera and add uniform noise to the depth. RGB samples of randomization on the \textit{Fetch Pick} environment can be seen in Figure ~\ref{fig:randomness}.

\section{RESULTS}

\begin{figure*}[!h]
\begin{center}
\includegraphics[width=6.5in]{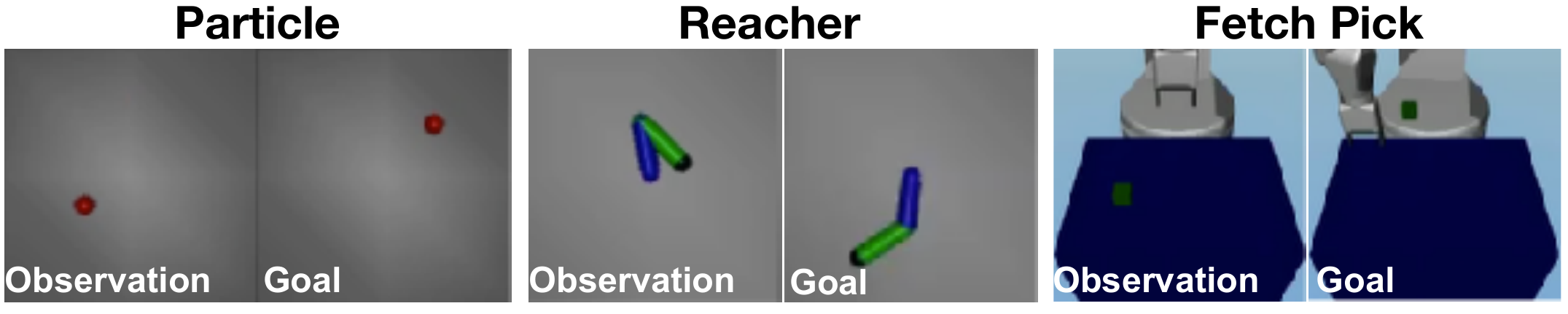}
\end{center}
\caption{To evaluate our method, we test on three different environments: ~\textit{Particle}, ~\textit{Reacher} and ~\textit{Fetch Pick}. Since we learn multi goal policies, the policy takes in both the observation at timestep $t$ and the desired goal for the episode.}
\label{fig:sim_envs}
\end{figure*}

To show the effectiveness of our method, we experiment on a range of simulated and real robot environments. In this section we first describe the environments. Following this, we discuss comparisons of our methods to baselines and show the utility of our method on improving training. Finally, we discuss real robot experiments. 
%%PA: from reading this, it could be interpreted that our key contribution is the bottlenecks, I think it should be rephrased to include utility of all our ideas

\subsection{Environments} \label{sec:environments}
Since there are no standard environments for multi-goal RL, we create three of our own simulated environments to test our method. The first two environments, \textit{Particle} and \textit{Reacher} are in a 2D workspace. 
%%PA: 2/3 being relatively makes the experiments sound a lot less exciting; can this phrasing be changed a bit? (while still accurate); also, are they really that easy from pixels?
The third environment \textit{Fetch Pick} is in a 3D workspace with a simulated version of the Fetch robot needing to pick up and place a block. All these environments are simulated in the \textit{MuJoCo}~\cite{todorov2012mujoco} physics simulator.

\vspace{0.1in}

\noindent (a) Particle: In this 2D environment the goal for the agent is to move the 2D particle to a given location. The state space is 4D and consists of the particle's location and velocity. The observation space is RGB images ($100\times100\times3$) from a camera placed above the scene. The action space is the 2D velocity of the particle. This action space allows for control on single RGB observations without requiring memory for velocity (since velocity cannot be inferred from a single RGB frame). The agent gets a sparse reward ($+1$) if the particle is within $\epsilon$ of the desired goal position and no reward ($0$) otherwise. The observation for the goal is an image of the particle in its desired goal position.

\vspace{0.1in}

\noindent (b) Reacher: In this 2D environment the goal for the agent is to move the end-effector of a two-link robot arm to a target location. The state space is 4D and consists of the joint positions and velocities. The observation space is RGB images ($100\times100\times3$) from a camera placed above the scene. The action space is the 2D velocities for the joints. The agent gets a sparse reward ($+1$) if the end-effector is within $\epsilon$ of the desired goal position and no reward ($0$) otherwise. The observation for the goal is an image of the reacher in its desired goal end-effector position. 

\vspace{0.1in}

\noindent (c) Fetch Pick: In this 3D environment with the simulated Fetch robot, the goal for the agent is to pick up the block on the table and move it to a given location in the air. The state space consists of the joint positions and velocities of the robot and the block on the table. The observation space is RGBD images ($100\times100\times4$) from a camera placed in front of the robot. The action space is 4D. Since this problem does not require gripper rotation, we keep it fixed. Three of the four dimensions of the action space specify the desired relative\footnote{The desired gripper position is relative to the current gripper position.} position for the gripper. The last dimension specifies the desired distance between the fingers of the gripper. The agent gets a sparse reward ($+1$) if the block is within $\epsilon$ of the desired goal block position. The observation for the goal is an image of the block in its desired goal block position and the Fetch arm in a random position. To make exploration in this task easier following~\cite{andrychowicz2017hindsight}, we record a \textit{single} state in which the box is grasped and start half of the training episodes from this state.

\begin{figure*}[!h]
\begin{center}
\includegraphics[width=7.0in]{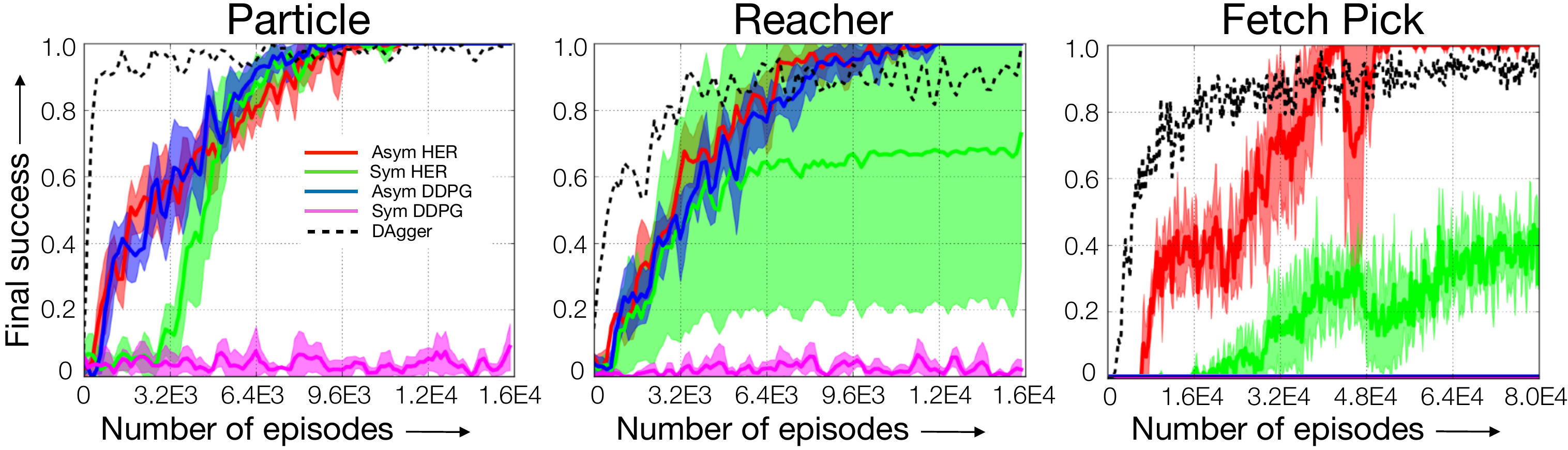}
\end{center}
\caption{We show that asymmetric inputs for training outperforms symmetric inputs by significant margins. The shaded region corresponds to $\pm 1$ standard deviation across 5 random seeds. Although the behaviour cloning (BC) by expert imitation baseline (dashed lines) learn faster initially, it saturates to a sub optimal value compared to asymmetric HER. Also note that the BC baseline doesn't include the iterations the expert policy was trained on.}
\label{fig:all_comp}
\end{figure*}

\begin{figure*}[!h]
\begin{center}
\includegraphics[width=7.0in]{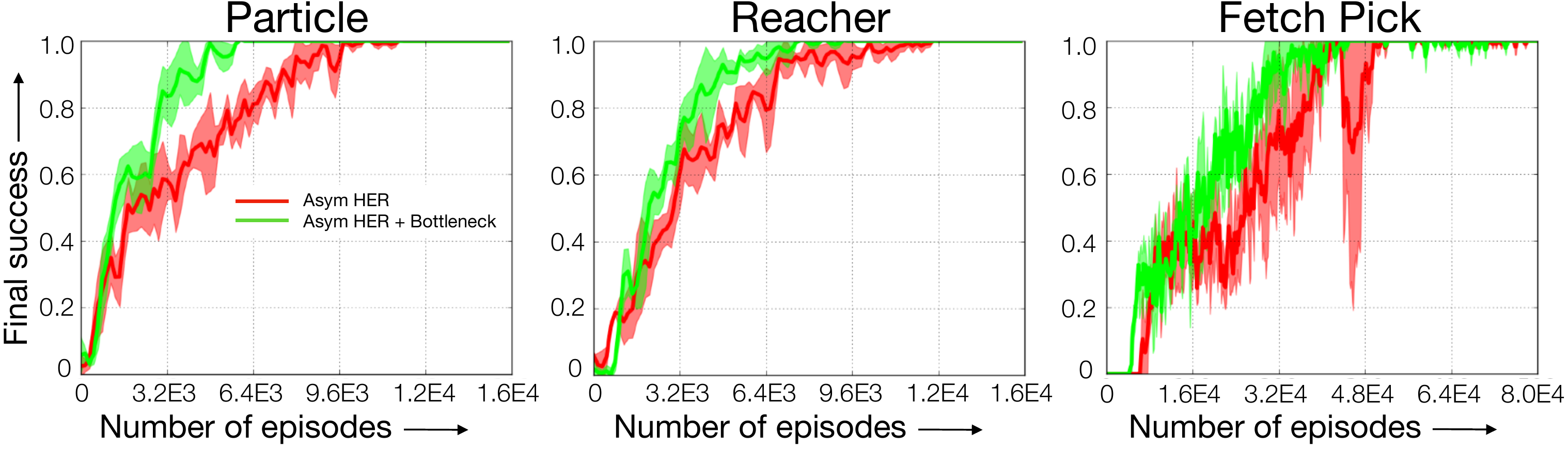}
\end{center}
\caption{We show that bottlenecks can be used to further improve training of our method. \textit{Particle} and \textit{Reacher}, the improvements are quite significant. On \textit{Fetch Pick}, we observe more stable training (lower variance denoted by shaded regions).}
\label{fig:bottleneck}
\end{figure*}

\subsection{Robot evaluation}
For our real world experiments we use a 7-DOF Fetch robotic arm\footnote{\url{fetchrobotics.com/platforms-research-development/}}, which is equipped with a two fingered parallel gripper. The camera observations for the real world experiments is an off the shelf Intel RealSense R200 camera that can provide aligned RGBD images. Since real depth often contains holes~\cite{oh2009hole}, we employ nearest neighbour hole filling to get better depth images~\cite{yang2012depth}. 
%%PA: can we cite something for the hole filling we do? or if that's in oh2009hole, move that citation to end of sentence
To further improve the depth, we cover/recolor parts of the robot that are black like parts of the torso and parts of the gripper.

\begin{figure*}[!h]
\begin{center}
\includegraphics[width=7.0in]{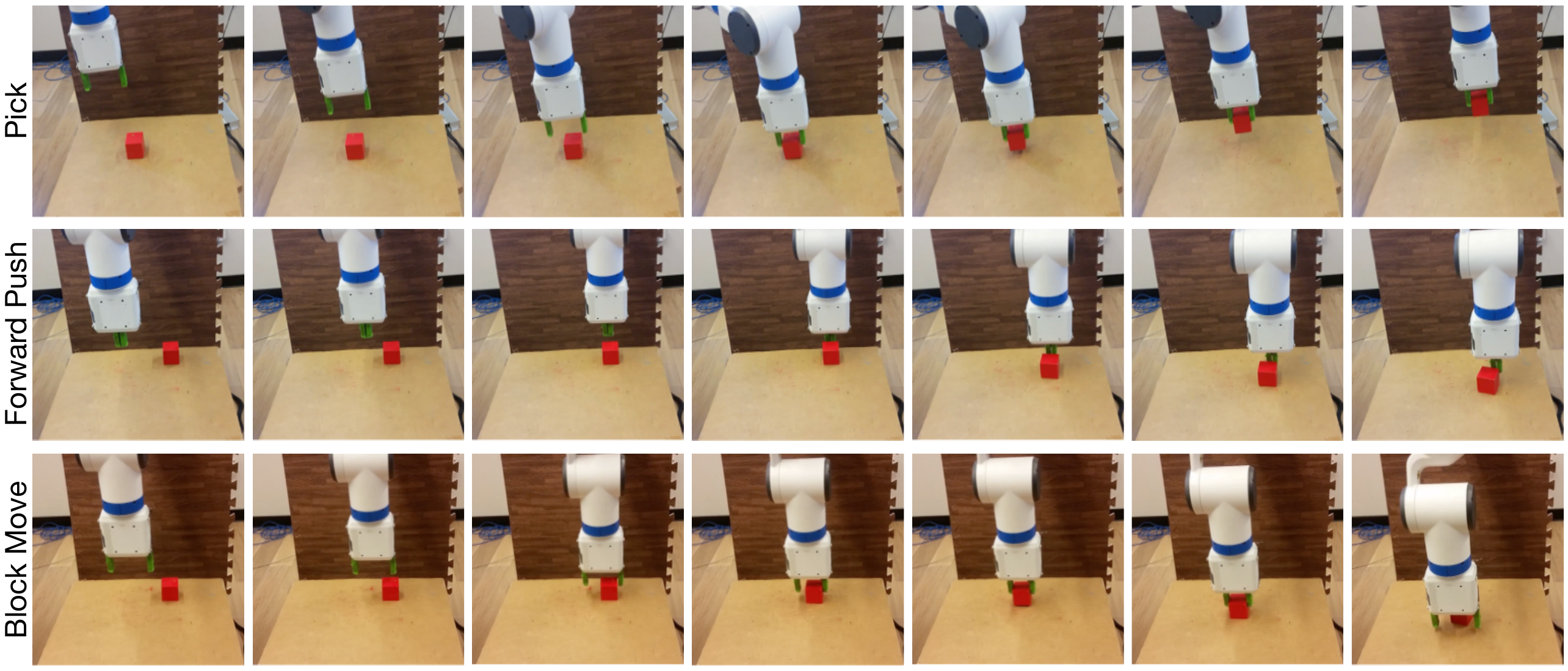}
\end{center}
\caption{Successive frames of our asymmetric HER policies on three real robot tasks show how our method can be successfully used for simulation to real transfer of complex policies. Full length experiments can be found in the \href{\URL}{videos} on \url{\URL}.}
\label{fig:real_robot}
\end{figure*}

We experiment on three tasks for the real robot. The first task is \textit{Pick} which is similar to the simulated task of \textit{Fetch Pick} described in Section ~\ref{sec:environments}(c). The second task is \textit{Forward Push}, where the robot needs to push the block forward\footnote{The fingers are blocked for this task to avoid grasping.}. The third task is \textit{Block Move}, where the robot needs to move the block to the target position on the table. In all tasks the goal is specified by an image of the box in the target location. A video of these experiments can be found in \url{\URL} and sample successes from our method in Figure ~\ref{fig:real_robot}.

The observations for the real robot tasks is an RGBD image from the physical camera placed in front of the robot. The goal observation for the actor is a simulated image describing the desired goal. We note that giving real world observations for the goal observation also works, however for consistency in evaluation, we use a simulated goal observation.

\subsection{Does asymmetric inputs to actor critic help?}
% \subsection{Comparisons to baselines}
%%PA: "comparison to baselines" doesn't sound super-strong, e.g. it doesn't sound like this is comparison to prior state of the art, rather against something quick and dirty to implement...
To study the effect of asymmetric inputs, we compare to the baseline of using symmetric inputs (images for both the actor and the critic networks). Figure~\ref{fig:all_comp} shows a summary of the final episodic rewards, with the x-axis being the number of episodes the agent experiences. As evident from the \textit{Particle} results, asymmetric input versions of both DDPG and HER perform much better than their symmetric counterparts. The simplicity of the \textit{Particle} may explain the similar performance between asymmetric DDPG and HER. \textit{Fetch Pick} is a much harder sparse reward task, which shows the importance of using HER over DDPG. In this case as well, the asymmetric version of HER performs significantly better than the symmetric version.

\subsection{Would imitation learning from an expert policy succeed?}
Imitation learning is a powerful technique in robotics~\cite{argall2009survey}. Hence a much stronger baseline is to behaviour clone from an expert policy. To do this we first train an expert policy~\cite{argall2009survey} on full states that performs the task perfectly. Now given this expert policy, we behaviour clone to a policy that takes the partial visual observations as input. We use DAgger~\cite{ross2011reduction} for better imitation/cloning.

Figure ~\ref{fig:all_comp} shows the final episodic rewards of the behaviour cloned policy (in dashed lines) with the x-axis being the number of demonstrations. As expected, the DAgger policy learns much faster initially (since it receives supervision from a much stronger expert policy). However in all the environments, it saturates in performance and is lower than than our method (asymmetric HER) for a large number of rollouts. One reason for this might be that behaviour cloning would fail if the expert policy depends on information contained in the full state but not in the partial observation.

\subsection{Can we speed up training?}
Another way of incorporating the full state from the simulator is by adding an auxiliary task of predicting the full state from partial observations. By adding a bottleneck layer~\cite{zhang2016vision} in the actor and adding an additional $L_2$ loss between the bottleneck output and the full state, we further speed up training. On our simulated tasks, these bottlenecks in the policy network improve the stability and speed of training (as seen in Figure~\ref{fig:bottleneck}).

%%PA: needs a bit of restructuring, because it's strange to have a "robot setup" in the middle of the experiments section as its own subsection

% \subsection{Robot setup}
% For our real world experiments we use a 7-DOF Fetch robotic arm\footnote{\url{fetchrobotics.com/platforms-research-development/}}, which is equipped with a two fingered parallel gripper. The camera observations for the real world experiments is an off the shelf Intel RealSense R200 camera that can provide aligned RGBD images. Since real depth often contains holes~\cite{oh2009hole}, we employ nearest neighbour hole filling to get better depth images~\cite{yang2012depth}. 
% %%PA: can we cite something for the hole filling we do? or if that's in oh2009hole, move that citation to end of sentence
% To further improve the depth, we cover/recolor parts of the robot that are black like parts of the torso and parts of the gripper.

% \subsection{Robot experiments}
% We experiment on three tasks for the real robot. The first task is \textit{Pick and Place} which is described in Section ~\ref{}. The second task is \textit{Forward Push}, where the robot needs to push the block forward\footnote{The fingers are blocked for this task to avoid grasping.}. The third task is \textit{Block Move}, where the robot needs to move the block to the target position on the table. In all tasks the goal is specified by an image of the box in the target location. A video of these experiments can be found in \URL and sample successes from our method in Figure ~\ref{fig:real_robot}.

\subsection{How well do these policies transfer?}
By combining asymmetric HER with domain randomization~\cite{tobin2017domain}, we show significant performance gains (see Table ~\ref{tab:real_robot}) compared to baselines previously mentioned. Our method succeeds on all the three tasks for all the 5 times the policy was run with different block initializations and goals. We also note that behaviours like \textit{push-grasping}~\cite{dogar2011framework} and \textit{re-grasping}~\cite{schlegl2001fast} emerge from these trained policies which can be seen in the \href{\URL}{video}. Among the baselines we evaluate against, we note that behaviour cloning with DAgger is the only one that performs reasonably (as seen in the \href{\URL}{video} and Table~\ref{tab:real_robot}).

\begin{table*}[h]
\centering
\caption{Comparison of asymmetric HER with baselines and ablations}
\label{tab:real_robot}
\begin{tabular}{c|c|cccc|cc}
\multicolumn{1}{l|}{} & \multirow{2}{*}{\textbf{Asym HER}} & \multicolumn{4}{c|}{Baselines} & \multicolumn{2}{c}{Randomization ablations} \\ \cline{3-8} 
 &  & Asym DDPG & Sym HER & Vanilla BC & DAgger BC & Without any & Without viewpoints \\ \hline
Pick & \textbf{5/5} & 0/5 & 0/5 & 0/5 & 3/5 & 0/5 & 0/5 \\
Forward Push & \textbf{5/5} & 0/5 & 0/5 & 1/5 & 0/5 & 0/5 & 0/5 \\
Block Move & \textbf{5/5} & 0/5 & 0/5 & 0/5 & 0/5 & 0/5 & 4/5
\end{tabular}
\end{table*}

\subsection{How important is domain randomization?}
To show the importance of randomization, we perform ablations (last two columns of Table ~\ref{tab:real_robot}) by training policies without any randomization and testing them in the real world. We notice that without any randomization, the policies fail to perform on the real robot while performing perfectly in the simulator. Another randomization ablation is by removing the viewpoint randomization while keeping the texture and lighting randomization during training. We notice that apart from the ~\textit{Block Move} task, removing viewpoint randomization severely affects performance.

\begin{figure}[!h]
\begin{center}
\includegraphics[width=3.2in]{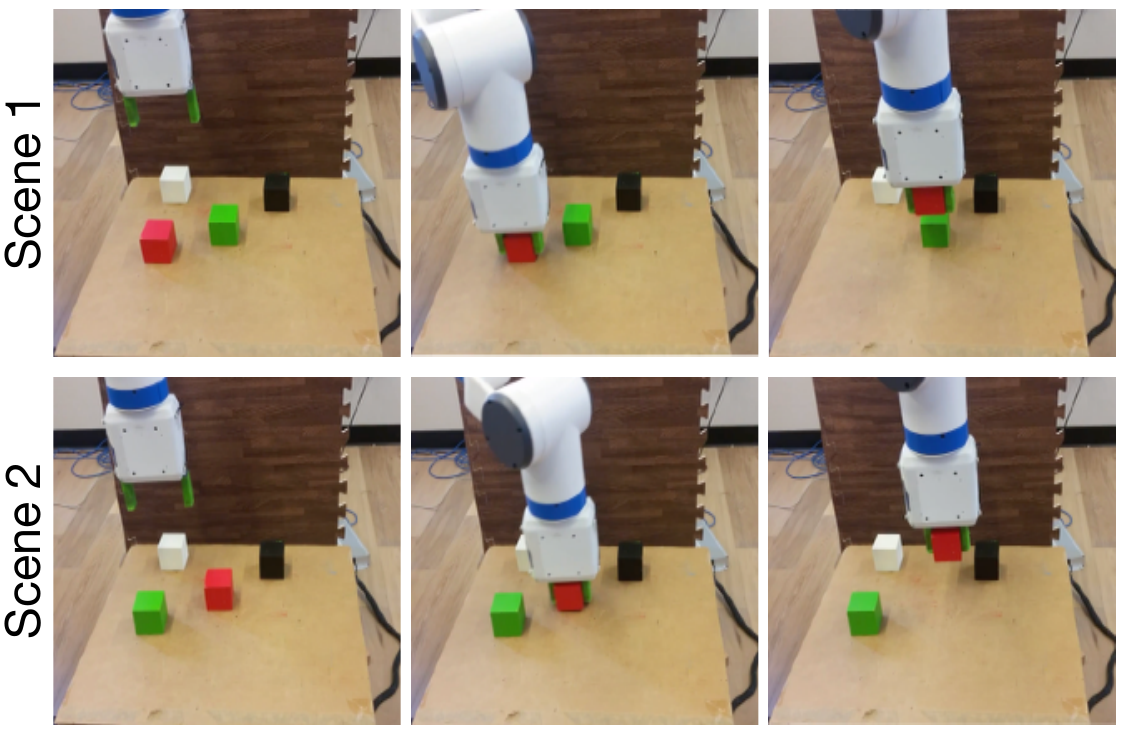}
\end{center}
\caption{Domain randomization during training allows the learned policies to be robust to distractors. Here we see how a policy trained to \textit{Pick} the red block is robust to distractor blocks. The difference in the two scenes shown here is that inspite of changing the initial location of the red block, the arm still picks the red block.}
\label{fig:robustness}
\end{figure}

Randomizing the observations in training also gives us an added benefit: robustness to distractors. Figure ~\ref{fig:robustness} shows the performance of our ~\textit{Pick} policy, which was trained on a single red block, work even in the presence of distractor blocks.

\subsection{Implementation Details}
In this section we provide more details on our training setup. The critic is a fully connected neural network with 3 hidden layers; each with 512 hidden units. The hidden layers use ReLU~\cite{krizhevsky2012imagenet} for the non linear activation. The input to the critic is the concatenation of the current state $s_t$, the desired goal state $g$ and the current action $a_t$. The actor is a convolutional neural network (CNN) with 4 convolutional layers with 64 filters each and kernel size of $2\times 2$. This network is applied on both the current observation $o_t$ and the goal observation $g^{o}$. The outputs of both CNNs are then concatenated and passed through a fully connected neural network with 3 hidden layers. Similar to the critic, the hidden layers have 512 hidden units each with ReLU activation. The output of this actor network is normalized by a tanh activation and rescaled to match the limits of the environment's action space. In order to prevent tanh saturation, we penalize the preactivations in the actor's cost.

During each iteration of DDPG, we sample 16 parallel rollouts of the actor. Following this we perform 40 optimization steps on minibatches of size 128 from the replay buffer of size $10^5$ transitions. The target actor and critic networks are updated every iteration with a polyak averaging of 0.98. We use Adam~\cite{kingma2014adam} optimization with a learning rate of 0.001 and the default Tensorflow~\cite{abadi2016tensorflow} values for the other hyperparameters. We use a discount factor of $\gamma=0.98$ and use a fixed horizon of $T=50$ steps. For efficient learning, we also normalize the input states by running averages of the means and standard deviations of encountered states.

The behavioural policy chosen for exploration chooses a uniform random action from the space of valid actions with probability $20\%$. For the rest $80\%$ probability, the output of the actor is added with coordinate independent Normal noise with standard deviation equal to $5\%$ of the action range.

\section{CONCLUSION}
In this work we introduce asymmetric actor-critic, a powerful way of utilizing the full state observability in a simulator. By training the critic on full states while training its actor on rendered images, we learn vision-based policies for complex manipulation tasks. Our extensive evaluation both in the simulator and on the real world robot shows significant improvements over standard actor-critic baselines. This method's performance is also superior to the much stronger imitation learning with DAgger baseline, even though it was trained without an expert. Coupled with domain randomization, our method is able to learn visual policies that works in the real world while being trained solely in a simulator. 

\bibliographystyle{IEEEtran}
\bibliography{IEEEabrv,references}
\end{document}